\documentclass[10pt,letterpaper]{article}

\usepackage{cogsci}
\usepackage[]{linguex}
\usepackage{times}
\usepackage{latexsym}
\usepackage{CJKutf8}
\usepackage{graphicx}
\usepackage{booktabs}

\makeatletter
\renewcommand{\@makefnmark}{\hbox{*}}
\makeatother

\cogscifinalcopy 

\usepackage[
  style=apa,
  natbib=true,
  annotation=false,
]{biblatex}
\usepackage{float} 

\title{How Do People Quantify Naturally: Evidence from Mandarin Picture Description}

 \author{
   {\large\bfseries Yayun Zhang (1360@mails.ccnu.edu.cn)$^1$, Guanyi Chen (g.chen@ccnu.edu.cn)$^1$\thanks{Corresponding Author}}, Fahime Same (fahimeh.same@gmail.com)$^2$, Saad Mahamood (s.mahamood@shopware.com)$^3$ \& Tingting He (tthe@ccnu.edu.cn)$^1$\\
   {\normalsize\normalfont
     $^1$Hubei Provincial Key Laboratory of Artificial Intelligence and Smart Learning, National Language Resources Monitoring and Research Center for Network Media, School of Computer Science, Central China Normal University \\
     $^2$Trivago N.V. $^3$Shopware
   }
 }

\begin{document}
\begin{CJK}{UTF8}{gbsn}
\maketitle

\begin{abstract}
Quantification is a fundamental component of everyday language use, yet little is known about how speakers decide whether and how to quantify in naturalistic production. We investigate quantification in Mandarin Chinese using a picture-based elicited description task in which speakers freely described scenes containing multiple objects, without explicit instructions to count or quantify. Across both spoken and written modalities, we examine three aspects of quantification: whether speakers choose to quantify at all, how precise their quantification is, and which quantificational strategies they adopt. Results show that object numerosity, animacy, and production modality systematically shape quantificational behaviour. In particular, increasing numerosity reduces both the likelihood and the precision of quantification, while animate referents and modality selectively modulate strategy choice. This study demonstrates how quantification can be examined under unconstrained production conditions and provides a naturalistic dataset for further analyses of quantity expression in language production.

\end{abstract}

\section{Introduction}







Quantification, the linguistic expression of \emph{how many} or \emph{how much}, is a pervasive component of everyday language use~\citep{moxey2023communicating}. When speakers describe scenes containing multiple objects, they are not obliged to encode quantity, nor to do so with a fixed level of precision. Instead, they flexibly decide whether to include any quantificational expression at all, how precise such expression should be, and which linguistic resources best serve communicative goals under perceptual and production constraints. Understanding quantification in this sense, therefore, requires examining it as a process of choice in language production, rather than solely as a matter of semantic interpretation.

A large body of research in cognitive science has established that quantity representation is closely tied to visual perception \citep{coventry2005grounding,coventry2010talking,Harvey2015-vw}. Humans can rapidly and accurately discriminate small numerosities through subitising, whereas larger quantities are represented approximately via the approximate number system~\citep{feigenson2004core,dakin2011common}. As magnitude increases, sensitivity to numerical differences decreases, yielding a graded notion of numerical precision that constrains how quantity can be linguistically encoded~\citep{haberman2009seeing,pezzelle-etal-2017-precise,PEZZELLE2018117}. 

\begin{figure}[t]
    \centering
    \includegraphics[width=0.75\linewidth]{img/7679.jpg}
    \caption{Example stimulus used in the elicited description task, depicting a scene with multiple deer.}
    \label{fig:example_animate}
\end{figure}

Languages exploit these perceptual representations through a range of quantificational mechanisms~\citep{szymanik2023questions}. Numeral-based expressions often support exact reading (e.g., \emph{five horses}), while other devices, such as vague or proportional quantifiers (e.g., \emph{many birds})
convey quantity in context-sensitive, approximate, or relational ways \citep{barwise1981generalized,kenney1996vagueness,coventry2010talking,cummins2012granularity,chen2023computational}. 
These mechanisms differ in their degree of numerical commitment and in how directly they map onto perceptual representations, enabling speakers to adapt quantification to context~\citep{moxey2023communicating,Ramotowska2024}. Importantly, this diversity of linguistic mechanisms creates a genuine decision space in production: speakers must choose not only \emph{what} quantity to express, but also \emph{how} to express it.

Despite extensive work on the semantics and interpretation of quantifiers, comparatively little is known about how speakers make these choices in naturalistic production~\citep{levelt1993speaking}. Much of the existing literature relies on comprehension, judgment, or acceptability tasks, which explicitly direct attention to quantity and constrain the set of available responses \citep{macuch2021,Nieuwland2010}. 
Such paradigms are well-suited for probing interpretation, but they obscure how quantification emerges when speakers are free to decide whether to quantify at all, and if so, with what level of precision and by which quantificational strategy. As a result, the production side of quantification in visually grounded unconstrained settings remains under-explored.

We examine quantification in Mandarin Chinese under naturalistic production conditions. Mandarin is a classifier language that provides multiple quantificational forms, yet relatively little work has examined how quantified expressions are produced in Chinese~\citep{zhan-levy-2018-comparing,Zhan2019AvailabilityBasedPP,chen-van-deemter-2022-understanding}. Quantity in Mandarin can be expressed through several mechanisms~\citep{keenan2012handbook}, including numeral–classifier noun phrases (e.g., 三个学生 ‘three students’), grouping and measure expressions that count containers, portions, or collective configurations (e.g., 两瓶水 ‘two bottles of water’, 一群鸟 ‘a flock of birds’), vague or proportional expressions (e.g., 很多 ‘many’, 大部分 ‘most’), and predicative constructions (e.g., 苹果很多 ‘apples are many’). This diversity makes Mandarin a useful case for examining how available quantificational resources are used in naturalistic, unconstrained production.

The present study investigates quantification in Mandarin Chinese using a picture-based elicited description task, with no explicit constraints on how scenes should be described. In this task, participants described images depicting multiple instances of the same object type (see Figure~\ref{fig:example_animate} for an example stimulus). By comparing spoken and written descriptions, we examine how production modality further modulates quantification choices. Specifically, we address three questions: (1) when describing multi-object scenes, do speakers choose to quantify at all; (2) when they do, how precise is their quantification; and (3) which quantificational strategies do they adopt under varying visual and conceptual conditions. By focusing on spontaneous production, this study aims to clarify how quantification arises from the interaction between visual perception, and production context in everyday language use.

\section{Research Questions and Hypotheses}

This study addresses three research questions concerning how Mandarin speakers quantify in naturalistic scene descriptions. Together, these questions capture a sequence of production decisions: whether speakers choose to quantify at all when describing multi-object scenes, how precise their quantification is when they do quantify, and which quantificational strategies they adopt under varying visual and conceptual conditions.

When describing a visual scene with multiple objects, speakers may or may not quantify. A description may omit any quantificational expression altogether (e.g., using merely a bare noun 苹果 `apples') or include a quantified expression of some kind (e.g., 许多苹果 `many apples'). Our first research question (RQ1), therefore, asks: \textbf{When describing multi-object scenes, do Mandarin speakers choose to quantify at all?} We treat the presence of quantification as a production choice that may be influenced by properties of the visual scene, characteristics of the referents, and the production modality, and, therefore, have the following hypotheses:
\begin{description}
    \item[Hypothesis 1 ($\mathcal{H}_1$)]
    As numerosity increases, sets become harder to represent as discrete collections. Larger numerosities are known to be represented approximately rather than exactly, which increases the cognitive cost of counting and committing to quantity in visually grounded contexts~\citep{feigenson2004core,dakin2011common}. We therefore hypothesise that \emph{increasing numerosity decreases the likelihood that speakers choose to quantify at all.}
    \item[Hypothesis 2 ($\mathcal{H}_2$)] Conceptual properties of referents may modulate the decision to quantify. Animate entities tend to be more salient and more readily construed as individuated sets than inanimate entities~\citep{grimm2012number}. We therefore hypothesise that \emph{speakers are more likely to quantify for animate referents than for inanimate referents}.
    \item[Hypothesis 3 ($\mathcal{H}_3$)] Production modality may further shape quantification decisions. Written production allows greater planning time and reduced time pressure compared to spontaneous speech, which can facilitate the inclusion of additional functional material~\citep{levelt1993speaking}. We therefore hypothesise that \emph{speakers are more likely to quantify in written descriptions than in spoken descriptions}.
\end{description}

When speakers do choose to quantify, the resulting expressions vary in their degree of precision. Quantification may be realised through exact numeral expressions (三个苹果 `three apples'), or vague/proportional expressions (许多苹果 `many apples'). Our second Research Question (RQ2) asks: \textbf{When speakers quantify, how precise is their quantification?} Drawing on work in numerical cognition and linguistic quantification, we hypothesise that precision is systematically related to numerosity and production modality: 
\begin{description}
    \item[Hypothesis 4 ($\mathcal{H}_4$)] As aforementioned, small numerosities are represented with higher precision than larger ones, whereas larger sets are encoded approximately~\citep{pezzelle-etal-2017-precise}. We hypothesise that smaller target sets favour precise quantification, while increasing numerosity leads speakers to shift toward less precise expressions.
     \item[Hypothesis 5 ($\mathcal{H}_5$)] Again, since written production supports more deliberate planning and monitoring than spoken production, we assume that \emph{written descriptions contain more precise quantification than spoken descriptions}.
\end{description}

When speakers choose to quantify, we distinguish three quantificational strategies in production: (1) The \textit{Exact} strategy uses numeral–classifier constructions to encode exact cardinality (e.g., 三个苹果 `three apples'); (2) The \textit{Grouping} strategy uses collective or measure classifiers to package multiple entities into a single unit (e.g., 一堆苹果 `a pile of apples'); (3) The \textit{Scalar} strategy relies on vague or proportional quantifiers that express quantity in an underspecified or context-dependent manner (e.g., 许多苹果 `many apples'). \footnote{Predicative constructions expressing quantity (e.g., 苹果很多 ‘apples are many’) occurred only rarely in our experiment. Given their low frequency, they were not analysed as a separate strategy but were grouped with the scalar strategy for the purposes of statistical analysis.}

The exact strategy constitutes an instance of precise quantification, whereas the grouping and scalar strategies instantiate vague quantification. Importantly, these two vague strategies realise vagueness in distinct ways: the Grouping strategy achieves vagueness by construing multiple entities as a collective unit, while the Scalar strategy encodes vagueness along a scalar dimension. This distinction allows us to examine not only when speakers engage in vague quantification, but also how vagueness is realised, through collective construal or through scalar imprecision.

Our third research question (RQ3) asks: \textbf{When Mandarin speakers quantify, which quantificational strategies do they adopt under varying visual and conceptual conditions?} We treat strategy choice as a categorical production decision reflecting how speakers exploit the available linguistic resources to express quantity.
\begin{description}
    \item[Hypothesis 6 ($\mathcal{H}_6$)] \emph{The factors influencing quantification and precision, i.e., numerosity, production modality, and animacy, also significantly modulate speakers’ choice of quantificational strategy.}
     \item[Hypothesis 7 ($\mathcal{H}_7$)] Animate entities are often construed as coherent groups, which licenses collective construals in quantified expression. We hypothesise that \emph{animate referents increase the likelihood of grouping strategies.}
\end{description}

\section{Experiment}

\subsection{Method}

\subsubsection{Material.}
The experimental stimuli consisted of 200 colour images showing various quantities of objects. Stimuli included both animate entities (e.g., birds, cows, sheep) and inanimate objects (e.g., apples, coffee beans, bottles), with 100 images in each category. The quantities shown in the images ranged from 2 to 99 items. Items were sampled from the VAQUUM dataset~\citep{wong-etal-2025-vaquum}. We selected the images manually to make sure that (1) the target objects were salient and clearly visible, (2) all objects within an image were of the same kind, (3) no other major object was present in the background, and (4) the number of objects was approximately uniformly distributed across the quantity range.  
\begin{figure}[t]
    \centering
    \includegraphics[width=0.6\linewidth]{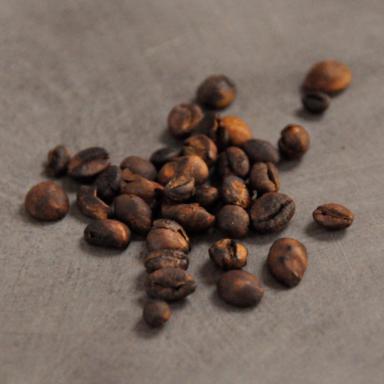}
    \caption{Example stimulus from the elicited description task, depicting multiple instances of an inanimate object (coffee beans).}
    \label{fig:example_inanimate}
\end{figure}

\subsubsection{Design.}
The experiment had a between-subject design, with modality (spoken vs. written) as the independent variable. Participants were randomly assigned to one of the two experiments. In the written experiment, participants saw the images on a monitor and typed their responses. In the spoken experiment, participants provided oral descriptions of the scenes that were audio-recorded. Within each modality, images were further divided into five groups, each containing 40 images.

\subsubsection{Procedure.}
Participants were tested individually in a quiet laboratory room. The experiment was conducted using a custom-built experimental interface developed in Python with the Tkinter library. After providing informed consent, the participants received instructions explaining that they would view a series of 40 images and should describe what they saw in each image. No specific constraints were placed on the content or length of descriptions. The entire experimental session lasted approximately 27.32 minutes per participant in the written experiment, and 20.07 minutes in the spoken experiment.

\subsubsection{Participants.}
Fifty participants participated in the experiment (mean age = 22.2 years, 37 female, 13 male), making each scene described by ten participants (5 written and 5 spoken). Participants were native speakers of Mandarin Chinese. They received 20 RMB as compensation for their participation. The experiment was conducted in person and in a laboratory setting. 

\subsection{Data Processing}

\begin{table*}[t]
\centering
\small
\caption{Example of elicited descriptions. The corresponding stimuli are shown in Figures~\ref{fig:example_animate} (animate) and~\ref{fig:example_inanimate} (inanimate). Quantified expressions analysed in this study are highlighted.}
\vskip 0.12in
\label{tab:example}
\begin{tabular}{llp{13cm}}
\toprule
Animacy & Modality & Example Description \\
\midrule
Animate & Written & \textcolor{blue}{12只动物在雪地里。}后面是挂满雪的树枝。\\
& & \emph{\textcolor{blue}{There are twelve animals in the snow.} Behind them are tree branches covered with snow.} \\
Animate & Spoken & 图片上应该是拍摄于冬天。\textcolor{blue}{然后有非常多只鹿。}然后它正在注视这边。鹿的上方是一些树枝。\\
& & \emph{The picture should have been taken in winter. \textcolor{blue}{And then there are a lot of deer.} And then one of them is looking this way. Above the deer are some tree branches.} \\
Inanimate & Written & \textcolor{blue}{大概有不到40粒咖啡豆在桌子上。}\\
& & \emph{\textcolor{blue}{There are probably fewer than forty coffee beans on the table.}}\\
Inanimate & Spoken & \textcolor{blue}{这张图片里面有很多咖啡豆。}咖啡豆的颜色是有点焦黄的。\\
& & \emph{\textcolor{blue}{There are many coffee beans in this image.} The colour of the coffee beans is kind of yellowish-brown.} \\
\bottomrule
\end{tabular}
\end{table*}

We collected a total of 2,000 descriptions for the 200 stimuli from 50 participants, including 1,000 written descriptions and 1,000 spoken descriptions. All spoken responses were transcribed into text using OpenAI’s Whisper\footnote{\url{https://github.com/openai/whisper}} model. Following transcription, all descriptions were manually reviewed to correct (1) typographical errors in the written responses and (2) transcription errors in the spoken responses. Throughout this process, discourse markers and fillers (e.g., 那个 `um/uh', 然后 `then', 呃 `uh/um') were retained, as they may be relevant for future analyses of discourse-level phenomena. Example descriptions corresponding to the stimuli shown in Figures~\ref{fig:example_animate} and~\ref{fig:example_inanimate} are provided in Table~\ref{tab:example}.

\subsubsection{Annotation.}


Each elicited description was annotated for the information relevant to the present analysis. As illustrated by the examples in Table~\ref{tab:example}, descriptions often include content unrelated to quantity, such as background information or details about individual objects. We therefore first identified and extracted the portion of each description that referred to the target objects of interest.

For each extracted expression—whether quantified or non-quantified—we annotated (1) the term for quantification, including quantifiers (e.g., 一些 ‘some’) and numeral–classifier constructions (e.g., 两个 ‘two CL’), etc.; (2) whether the expression instantiated vague quantification; and (3) the quantificational strategy (i.e., exact, grouping, or scalar). All annotations were performed by two independent annotators. In cases of disagreement, differences were resolved through discussion until consensus was reached.

\section{Results}

\begin{table*}[t]
\centering
\small
\caption{Top quantification terms produced under each combination of animacy (animate vs. inanimate) and modality (written vs. spoken). Terms are ranked by frequency within each condition; raw counts are shown in parentheses. NONE indicates descriptions in which no quantificational expression was produced.}
\vskip 0.12in
\label{tab:top_term}
\begin{tabular}{llp{14cm}}
\toprule
Animacy & Modality & Top-10 Quantification Terms\\
\midrule
Animate & Written &  一群 (187; `a flock of '), NONE (70), 很多 (48; `a lot of'), 许多 (19; `many'), 很多只 (16; `a lot of CL'), 鸟群 (13; `flock'), 成群 (10; `swarms of'), 一些 (7; `some'), 一大群 (6; `a large flock of'), 两只 (5; `two CL') \\
Animate & Spoken & 一群 (121; `a flock of '), 很多 (69; `a lot of'), NONE (58), 许多 (43; `many'), 很多只 (21; `a lof of CL'), 一些 (21; `some'), 非常多只 (15; `a great many CL'), 一堆 (12; `a pile of'), 许多只 (10; `many'), 非常多 (6; `a great many') \\
Inanimate & Written & NONE (119), 很多 (59; `a lot of'), 一堆 (46; `a pile of'), 一串 (21; `a string of'), 许多 (13; `many'), 一些 (12; `some'), 一盘 (7; `a plate of'), 两串 (5; `two strings of'), 几十个 (5; `dozens of'), 很多个 (5; `many CL')\\
Inanimate & Spoken & NONE (85), 很多 (52; `a lot of'), 一堆 (33; `a pile of'), 许多 (25; `many'), 一些 (24; `some'), 一串 (20; `a string of'), 非常多 (17; `a great many'), 很多个 (16; `many'), 各式各样 (7; `all kinds of'), 很多颗 (6; `many') \\
\bottomrule
\end{tabular}
\end{table*}

\begin{figure}[t]
    \centering
    \includegraphics[width=0.9\linewidth]{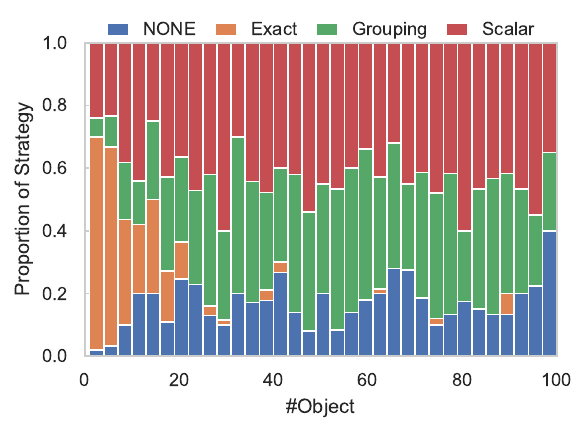}
    \caption{Distribution of quantificational strategies, including non-quantified descriptions, across object numerosities. Numerosities are grouped into bins of four. \emph{NONE} denotes non-quantified descriptions using bare nouns.}
    \label{fig:dist}
\end{figure}

Before turning to statistical modelling to address our research questions, we first present descriptive statistics summarising the use of quantificational terms and strategies in the dataset. Table~\ref{tab:top_term} reports the most frequent quantification terms produced under each combination of animacy and modality.
Across conditions, a small set of highly frequent terms accounts for a substantial proportion of descriptions. While both vague quantifiers and grouping expressions instantiate vague quantification, their relative prominence differs across conditions: grouping expressions such as 一群 ‘a flock of’ are particularly frequent for animate referents, whereas vague quantifiers are more evenly distributed across animacy and modality. Notably, descriptions without explicit quantification (NONE) are common across all conditions.

Figure~\ref{fig:dist} complements the above observations by showing the distribution of quantificational strategies, including non-quantified descriptions, across object numerosities, revealing systematic shifts in strategy use as set size increases. The exact strategy—realised through numeral–classifier constructions and instantiating precise quantification—is largely confined to smaller set sizes and becomes increasingly rare as numerosity increases. As set size grows, the decline of exact strategies is not primarily compensated by scalar strategies, but instead by increased use of the grouping strategy, which construes multiple entities as a collective unit, even though both grouping and scalar strategies instantiate vague quantification. Descriptions without explicit quantification (i.e., NONE) also account for a substantial proportion of responses at larger numerosities.

\subsection{When do Chinese Speakers Quantify?}

To address the first research question, whether speakers choose to quantify at all when describing multi-object scenes, we modelled the presence versus absence of quantification as a binary outcome. Specifically, we fitted a Logistic Mixed-Effects Model with \textsc{Quantified} (1 = an explicit quantificational expression was produced; 0 = no quantification) as the dependent variable.

The fixed effects were selected to reflect the hypotheses associated with this research question. \textsc{Object Numerosity} was included to test the hypothesis that increasing set size reduces the likelihood of quantification ($\mathcal{H}_1$). \textsc{Animacy} (animate vs. inanimate) was included to capture potential differences in how readily speakers quantify animate versus inanimate referents ($\mathcal{H}_2$). \textsc{Production Modality} (written vs. spoken) was included to assess whether the likelihood of quantification differs between written and spoken descriptions ($\mathcal{H}_3$). To account for repeated observations from the same participants and the same visual stimuli, the model included random intercepts for \textsc{Subject} and \textsc{Picture}. We computed p-values for fixed effects based on Wald z approximations.

\begin{table}[t]
\centering
\footnotesize
\caption{Logistic mixed-effects model predicting whether a description contained an explicit quantificational expression. Odds ratios (OR) are reported for ease of interpretation. `ani.' and `spo.' stand for `animate' and `spoken', respectively.}
\vskip 0.12in
\label{tab:rq1_logistic}
\begin{tabular}{lrrrrr}
\toprule
& Estimate & SE & $z$ & $p$ & OR \\
\midrule
Intercept & 2.66 & 0.07 & 39.28 & $< .001$ & 14.28 \\
\#Object & $-0.30$ & 0.02 & $-16.64$ & $< .001$ & 0.74 \\
Animacy (ani.) & 0.75 & 0.10 & 7.12 & $< .001$ & 2.11 \\
Modality (spo.) & 0.46 & 0.10 & 4.60 & $< .001$ & 1.59 \\
\bottomrule
\end{tabular}
\end{table}

Results indicate that all three predictors significantly modulate whether speakers choose to quantify (Table~\ref{tab:rq1_logistic}). Object numerosity showed a strong negative effect: a one-unit increase in log-transformed object count was associated with a substantial decrease in the likelihood of quantification (Est.=$-0.30$, OR=0.74). This pattern is in line with the descriptive distribution shown in Figure~\ref{fig:dist}, where non-quantified descriptions become increasingly prevalent as object numerosity increases.

Animacy also had a robust effect, with descriptions of animate referents being more than twice as likely to include explicit quantification compared to inanimate referents (Est. = 0.75, OR = 2.11). In addition, production modality significantly influenced quantification, such that spoken descriptions were more likely to contain explicit quantity expressions than written descriptions (Est. = 0.46, OR = 1.59), contrary to $\mathcal{H}_3$. Together, these results show that the decision to quantify is jointly shaped by numerosity, animacy, and production modality.

\subsection{How Precise are Chinese Speakers?}

Building on the analysis of RQ1, we next examined how precise quantification is when speakers choose to quantify (RQ2). We modelled \textsc{Precision} as a binary outcome using a Logistic Mixed-Effects Model, coding whether a quantified description used an exact numeral–classifier construction (1) or a non-exact expression (0). The analysis was restricted to descriptions containing an explicit quantificational expression.

The fixed effects were selected to test the hypotheses associated with RQ2. \textsc{Object Numerosity} was included to assess whether increasing set size reduces the likelihood of exact quantification ($\mathcal{H}_4$). \textsc{Production Modality} was included to test whether written descriptions favour more precise quantification than spoken descriptions ($\mathcal{H}_5$). As in the RQ1 analysis, random intercepts for \textsc{Subject} and \textsc{Picture} were included to account for repeated observations.

\begin{table}[t]
\centering
\footnotesize
\caption{Logistic mixed-effects model predicting the precision of quantification in quantified descriptions. Precision was coded as exact (1) versus non-exact (0). }
\vskip 0.12in
\label{tab:rq2_logistic}
\begin{tabular}{lrrrrr}
\toprule
& Estimate & SE & $z$ & $p$ & OR \\
\midrule
Intercept & 5.18 & 0.13 & 39.46 & $< .001$ & 178.11 \\
\#Object & $-2.75$ & 0.05 & $-59.11$ & $< .001$ & 0.06 \\
Modality (spo.) & $-1.37$ & 0.21 & $-6.59$ & $< .001$ & 0.25 \\
\bottomrule
\end{tabular}
\end{table}

Results show that both object numerosity and production modality significantly modulate the precision of quantification (Table~\ref{tab:rq2_logistic}). Object numerosity had a strong negative effect on precision: as the number of objects increased, speakers were substantially less likely to use exact numeral–classifier constructions rather than non-exact expressions (Est.=$-2.75$, OR=0.06). This sharp decline in precise quantification is again in line with the descriptive patterns shown in Figure~\ref{fig:dist}, where precise expressions are largely confined to smaller set sizes and rapidly give way to alternative strategies as numerosity increases.

Production modality also significantly affected quantificational precision. Compared to written descriptions, spoken descriptions were less likely to employ precise quantification (Est.=$-1.37$, OR=0.25).

\subsection{How do Chinese Speakers Quantify?}

\begin{table*}[t]
\centering
\footnotesize
\caption{Results of three Logistic Mixed-Effects Models predicting the use of different quantificational strategies.}
\vskip 0.12in
\label{tab:rq3_strategies}
\begin{tabular}{lrrrrrrrrr} 
\toprule
& \multicolumn{3}{c}{Numeral-Classifier} 
& \multicolumn{3}{c}{Grouping} 
& \multicolumn{3}{c}{Vague} \\ \cmidrule(lr){2-4} \cmidrule(lr){5-7} \cmidrule(lr){8-10}
& Est. (SE), $z$ & $p$ & OR
& Est. (SE), $z$ & $p$ & OR
& Est. (SE), $z$ & $p$ & OR \\
\midrule

& $-3.58$ (0.13), $-27.2$ & $<.001$ & 0.03
& $-0.85$ (0.07), $-12.9$ & $<.001$ & 0.43
& $-0.19$ (0.06), $-3.0$ & $.003$ & 0.83 \\

\#Object
& $-2.43$ (0.09), $-27.7$ & $<.001$ & 0.09
& $0.66$ (0.07), $9.0$ & $<.001$ & 1.93
& $0.40$ (0.06), $6.3$ & $<.001$ & 1.50 \\

Animacy (ani.)
& $-1.63$ (0.22), $-7.5$ & $<.001$ & 0.20
& $1.21$ (0.09), $13.7$ & $<.001$ & 3.35
& $-0.63$ (0.09), $-7.4$ & $<.001$ & 0.53 \\

Modality (spo.)
& $-1.60$ (0.21), $-7.6$ & $<.001$ & 0.20
& $-1.03$ (0.09), $-10.9$ & $<.001$ & 0.36
& $1.22$ (0.09), $13.9$ & $<.001$ & 3.40 \\

\bottomrule
\end{tabular}
\end{table*}

We next examined which quantificational strategies speakers adopt when producing quantified descriptions (RQ3). We modelled strategy choice using three separate Logistic Mixed-Effects Models, one for each quantificational strategy. In each model, the dependent variable coded whether a given strategy was used (1) or not (0), relative to the other two strategies. 

The fixed effects in all three models included \textsc{Object Numerosity}, \textsc{Production Modality}, and \textsc{Animacy}, reflecting the hypotheses associated with RQ3 that the factors influencing whether speakers quantify and how precisely they do so also shape the choice of quantificational strategy. As in the previous analyses, random intercepts for \textsc{Subject} and \textsc{Picture} were again included.

Consistent with $\mathcal{H}_6$, all three factors—object numerosity, production modality, and animacy—significantly modulated speakers’ choice of quantificational strategy. Increasing numerosity strongly reduced the likelihood of the exact strategy (Est.=$-2.43$, OR=0.09), while simultaneously increasing the likelihood of both the grouping strategy (Est.=0.66, OR=1.93) and the scalar strategy (Est.=0.40, OR=1.50). Importantly, the increase in vague quantification strategies was not uniform: grouping strategies showed a steeper increase than scalar strategies, suggesting that speakers preferentially shift toward grouping-based construals rather than scalar vagueness as set size grows.

Production modality also exerted systematic effects on strategy selection. Written descriptions were more likely than spoken descriptions to employ the exact strategy (Est.=$-1.60$, OR=0.20) and the grouping strategy (Est. = $-1.03$, OR = 0.36), whereas spoken descriptions were substantially more likely to use the scalar strategy (Est.=1.22, OR=3.40). Thus, within quantified descriptions, spoken production favours scalar vagueness, whereas written production favours both precision and collective construal.

Turning to $\mathcal{H}_7$, animacy had a strong and selective effect on the grouping strategy. Animate referents substantially increased the likelihood of using the grouping strategy (Est.=1.21, OR=3.35), while decreasing the likelihood of both the exact strategy (Est.=$-1.63$, OR=0.20) and the scalar strategy (Est.=$-0.63$, OR=0.53). This pattern supports the hypothesis that animate entities are preferentially construed as coherent groups, thereby licensing grouping-based quantificational strategies over alternative realisations of vagueness.


\section{Discussion}

The present study investigated how Mandarin speakers express quantity in a naturalistic picture description task, focusing on whether speakers quantify at all, how precise their quantification is, and which quantificational strategies they adopt. Across all analyses, object numerosity, animacy, and production modality systematically shaped quantificational behaviour, but their effects were distributed unevenly across these aspects.

Object numerosity emerged as a dominant factor. As set size increased, speakers were less likely to include explicit quantification, and when they did quantify, they were substantially less likely to use the exact strategy. This pattern is consistent with well-established limits on perceptual discriminability at higher numerosities, and extends prior findings by showing that numerosity also affects the decision to quantify at all in unconstrained production contexts.

Importantly, increasing numerosity did not simply increase scalar strategies, but instead favoured the grouping strategy, indicating that non-exact quantification is not homogeneous. The preferential shift toward grouping at higher numerosities suggests that speakers adapt their construal of quantity in response to perceptual and communicative pressures.

Animacy showed a more selective pattern of effects. While animate referents were more likely to be quantified overall, animacy primarily modulated strategy choice: animate entities strongly favoured the grouping strategy while disfavouring both the exact strategy and the scalar strategy. This asymmetry is consistent with the idea that animate entities are readily construed as coherent collectives, making the grouping strategy a natural option for expressing quantity in such cases.

Production modality further modulated how quantification was realised. Spoken descriptions were more likely to include explicit quantification overall, but when speakers quantified, written descriptions favoured the exact and grouping strategies, whereas spoken descriptions relied more on scalar strategies. Within non-exact quantification, spoken modality thus favoured scalar vagueness over collective construal, indicating that modality shapes not only whether speakers quantify, but also how vagueness is realised.

Overall, these findings underscore the value of examining quantification in naturalistic production. By allowing speakers to describe visual scenes freely, the present study reveals how perceptual constraints, conceptual properties of referents, and production context jointly shape quantificational choices. 

\section{Conclusion}

This study examined how Mandarin speakers express quantity in naturalistic scene descriptions, focusing on whether speakers choose to quantify, how precise their quantification is, and which quantificational strategies they adopt. We show that object numerosity, animacy, and production modality systematically shape quantificational behaviour, and that vague quantification in production is not uniform, but distinguishes grouping-based strategies from vague scalar expressions. We hope the elicited corpus in this study provides a resource for future analyses of quantification in language production, including cross-linguistic, discourse-level, and model-based investigations.

\section{Acknowledgment}

This work was supported by the National Language and Character Research Base and the MOE (Ministry of Education in China) Project of Humanities and Social Sciences (Project No.25YJC740005).

\paragraph{Use of AI Assistants.} In this work, we used GPT-5.2 to assist with refining the language of the paper and to help write code for data preprocessing, result analysis, and figure drawing. All experimental design decisions, analyses, and interpretations were made by the authors.


\printbibliography

\end{CJK}
\end{document}